\documentclass[letterpaper, 10 pt, conference]{ieeeconf}  

\IEEEoverridecommandlockouts                                      
\usepackage{algorithm}
\usepackage{algpseudocode}
\usepackage{amsmath}
\usepackage{amssymb}
\usepackage{booktabs}
\usepackage{graphicx}
\graphicspath{image/}
\usepackage{pdfpages}
\usepackage[font=small]{caption}
\usepackage{color, colortbl}
\usepackage{hyperref}
\usepackage{makecell}
\usepackage{caption}
\usepackage[capitalise,noabbrev]{cleveref}
\Crefname{figure}{\text{Fig.}}{\text{Figs.}}	
\Crefname{figure}{Figure}{Figures}  
\Crefname{equation}{\text{Eq.}}{\text{Eqs.}}

\title{\LARGE \bf
Controllable Latent Diffusion for Traffic Simulation
}
\author{\authorblockN{
        Yizhuo Xiao,
        Mustafa Suphi Erden,
        Cheng Wang}\thanks{All authors are with the School of Engineering and Physical Sciences, Heriot-Watt University, EH14 4AS, UK. Corresponding author: cheng.wang@hw.ac.uk. This work was funded by UK Research and Innovation (UKRI) under the UK government’s Horizon Europe funding guarantee [grant number EP/Z533464/1].}
}

\begin{document}

\maketitle

\thispagestyle{empty}
\pagestyle{empty}

\begin{abstract}
The validation of autonomous driving systems benefits greatly from the ability to generate scenarios that are both realistic and precisely controllable. Conventional approaches, such as real-world test drives, are not only expensive but also lack the flexibility to capture targeted edge cases for thorough evaluation. To address these challenges, we propose a controllable latent diffusion that guides the training of diffusion models via reinforcement learning to automatically generate a diverse and controllable set of driving scenarios for virtual testing. Our approach removes the reliance on large-scale real-world data by generating complex scenarios whose properties can be finely tuned to challenge and assess autonomous vehicle systems. Experimental results show that our approach has the lowest collision rate of \(0.226\) and lowest off-road rate of \(0.136\), demonstrating superiority over existing baselines. The proposed approach significantly improves the realism, stability and controllability of the generated scenarios, enabling more nuanced safety evaluation of autonomous vehicles.

\end{abstract}


\section{Introduction}
As Autonomous Vehicles (AVs) become increasingly integrated into modern transportation systems, ensuring their safe operation has emerged as a critical concern for both industry and society at large \cite{wang2021online}. However, the deployment of AVs on public roads necessitates rigorous validation of their safety across a wide array of driving conditions. Comprehensive testing is essential to validate that AVs can handle not only common driving scenarios but also unexpected, high-risk situations that may arise \cite{kalra2016driving}.

Current simulation techniques employed to assess AV performance include deterministic replay of real-world scenarios and rule-based simulations \cite{ding2023survey}. Deterministic replay reproduces scenarios recorded from driving logs, allowing AVs to be tested against situations they are likely to encounter. Rule-based simulations, in contrast, use predefined rules and models to generate driving scenarios in a controlled environment for testing specific behaviors. While these methods are effective for evaluating standard driving conditions and many known challenges—including the long-tail effect, where rare, high-risk scenarios (such as complex interactions with erratic drivers, sudden obstacles, adverse weather conditions, or unexpected vehicle malfunctions) occur infrequently—they still struggle to systematically generate a sufficient number of these critical long-tail scenarios for robust safety validation.

The limitation of existing datasets and simulation environments lies in their inability to capture the vast diversity and criticality needed to evaluate AVs across extreme conditions \cite{feng2020testing}. Scenarios involving rare events are underrepresented, making it difficult to assess an AV’s performance in situations that could lead to accidents. Addressing this gap requires advanced scenario generation techniques capable of producing a wide range of challenging, high-risk scenarios. Such techniques should not only replicate known dangerous situations but also generate novel scenarios that test the limits of an AV’s decision-making capabilities.

Recent advancements in generative artificial intelligence (GAI) offer promising solutions to this challenge. Generative models, particularly diffusion models (DMs) \cite{ho2020denoising}, have shown remarkable success in creating complex and high-dimensional data distributions \cite{song2019generative}. DMs work by learning to reverse a gradual noising process, effectively generating data samples that are statistically similar to the training data. Their ability to model intricate patterns makes them well-suited for generating realistic and diverse driving scenarios that include rare and safety-critical events.

Integrating DMs with reinforcement learning (RL) \cite{sutton1998reinforcement} further enhances their capability to generate meaningful and challenging scenarios \cite{lee2023aligning}. RL introduces a feedback mechanism where the generative model receives rewards or penalties based on the safety-critical nature of the generated scenarios. This combination allows the model to learn not just from the data distribution but also from the specific requirements of safety testing, encouraging the generation of scenarios that are most valuable for evaluating AV performance. However, existing RL-guided sampling in DM has limited generalization and can have unstable and unintended artifacts in the generated outputs.

In this work, we propose the controllable latent diffusion (CLD)\footnote{{\bf CLD code and video:} \url{https://github.com/RoboSafe-Lab/Controllable-Latent-Diffusion-for-Traffic-Simulation} \label{fn1}} that decomposes training into two distinct phases. In the first phase, we train an autoencoder to map high-dimensional driving trajectories into a compact latent space, where a DM is subsequently trained. This approach significantly reduces computational overhead while preserving the essential features of the data. In the second phase, the latent states generated by the denoising process are decoded and combined with a reward module to further guide the training of the DM. By incorporating a simple reward mechanism, our method conditions the DM to generate trajectories that meet user-specified controllability criteria. Our contributions are threefold:
\begin{itemize}
    \item We treat the denoising process as a Markov Decision Process (MDP) and incorporate RL to steer the denoising process by adjusting the DM gradient, enabling inherent optimized DM that generates constraint-compliant trajectories.
    \item We leverage important sampling and design a flexible reward-guided training strategy for DMs, significantly enhancing iteration efficiency and enabling fine-grained, user-specified control over the generated trajectories.
    \item Extensive experiments demonstrate that CLD not only captures the intrinsic data distribution to achieve high realism but also enforces controllability, producing diverse and adaptable traffic scenarios for AV testing.
\end{itemize}
The rest of the paper is organized as follows. \cref{sec:2} reviews related work on scenario generation, covering various approaches employed in current research. In \cref{sec:3}, we describe our proposed methodology in detail, outlining the overall pipeline and defining the components at each stage. Finally, \cref{sec:4} presents our experimental evaluation, where we compare our results against several baseline methods.

\section{Related Work}
\label{sec:2}
In this section, we present three popular ways to generate traffic scenarios: \textit{RL-based}, \textit{guided generative models-based} and \textit{optimization-based} methods.

\subsection{RL-based scenario generation}
\label{sec:2.1}
Goal-conditioned RL \cite{andrychowicz2017hindsight}, \cite{kaelbling1993learning}, \cite{nachum2018data}, \cite{schaul2015universal} is a powerful approach aimed at training agents capable of reaching specified goals by predicting action sequences that lead to the desired states. For instance, \cite{ransiek2024goose} applies this method to generate safety-critical scenarios by defining specific goals that agents use in creating challenging test cases for AVs. Similarly, \cite{koren2018adaptive} formulates the testing process as a MDP and leverages RL techniques to identify failure scenarios that expose vulnerabilities in AVs. 
However, these RL-based methods are not inherently generative and thus tend to produce a limited variety of scenarios. In contrast, our approach integrates RL within a DM-based generative framework, enabling the synthesis of a diverse set of realistic and controllable traffic scenarios, which is crucial for comprehensive AV testing.

\subsection{Guided generative models for scenario generation} 
\label{sec:2.2}
Generative models have demonstrated remarkable success in creating diverse and realistic scenarios for autonomous systems. These models effectively capture the complexities of real-world environments, making them a powerful tool for scenario generation. Generative adversarial networks (GANs) \cite{goodfellow2020generative} generate highly realistic samples by leveraging a competitive training process between a generator and a discriminator. Variational Autoencoders (VAEs) \cite{kingma2013auto} offer a structured latent space that facilitates controlled scenario manipulation. DMs, on the other hand, excel at generating high-quality, multi-modal distributions while maintaining stability during training. 

To achieve specific objectives in scenario generation, these generative models are often guided through various methodologies. For instance, in \cite{ajay2022conditional, chen2023playfusion, chen2023executing, chi2023diffusion, dabral2023mofusion, he2023diffusion, li2024crossway, black2023training}, the sampling phase of DMs is guided to satisfy the defined requirements for generating scenarios. Despite the success of guided generative models, many existing approaches operate directly in high-dimensional spaces, incurring substantial computational overhead. This inefficiency hinders the exploration and generation of complex scenarios, often at the expense of fidelity to real-world dynamics. Additionally, the inference process is guided without directly optimizing the model, which may result in suboptimal solutions.

\subsection{Optimization-based scenario generation}
\label{sec:2.3}
Evolutionary algorithms have been widely used to generate challenging test scenarios for AV by optimizing scenario parameters to identify the most critical and difficult situations for evaluation \cite{klischat2019generating}. Similarly, optimization techniques have been applied to adjust the initial states of traffic participants, progressively reducing the feasible solution space for collision avoidance and thus creating safety-critical scenarios for the rigorous testing of AV systems \cite{althoff2018automatic}. Building on this approach, \cite{rocklage2017automated} leverages optimization methods to automatically create diverse and high-stakes scenarios, focusing on uncovering system vulnerabilities during regression testing and ensuring robustness across software iterations.
Similarly, \cite{klischat2020scenario} introduces a workflow that extracts diverse road networks from OpenStreetMap\footnote{{\bf OpenStreetMap:} \url{https://www.openstreetmap.org} \label{fn2}} and uses the SUMO \cite{behrisch2011sumo} traffic simulator to generate traffic scenarios. By employing nonlinear optimization, it identifies safety-critical situations that can challenge autonomous systems, providing a scalable approach to scenario generation without relying on real-world data. 

However, optimization-based methods face several drawbacks. They rely heavily on iterative searches and heuristic algorithms, often resulting in high computational costs and slow convergence. Moreover, these techniques require carefully designed objective functions and generally struggle to scale with high-dimensional data, limiting their ability to capture the full diversity and complexity of real-world driving scenarios.

\begin{figure*}[htbp]
\centering
\includegraphics[width=0.9\textwidth]{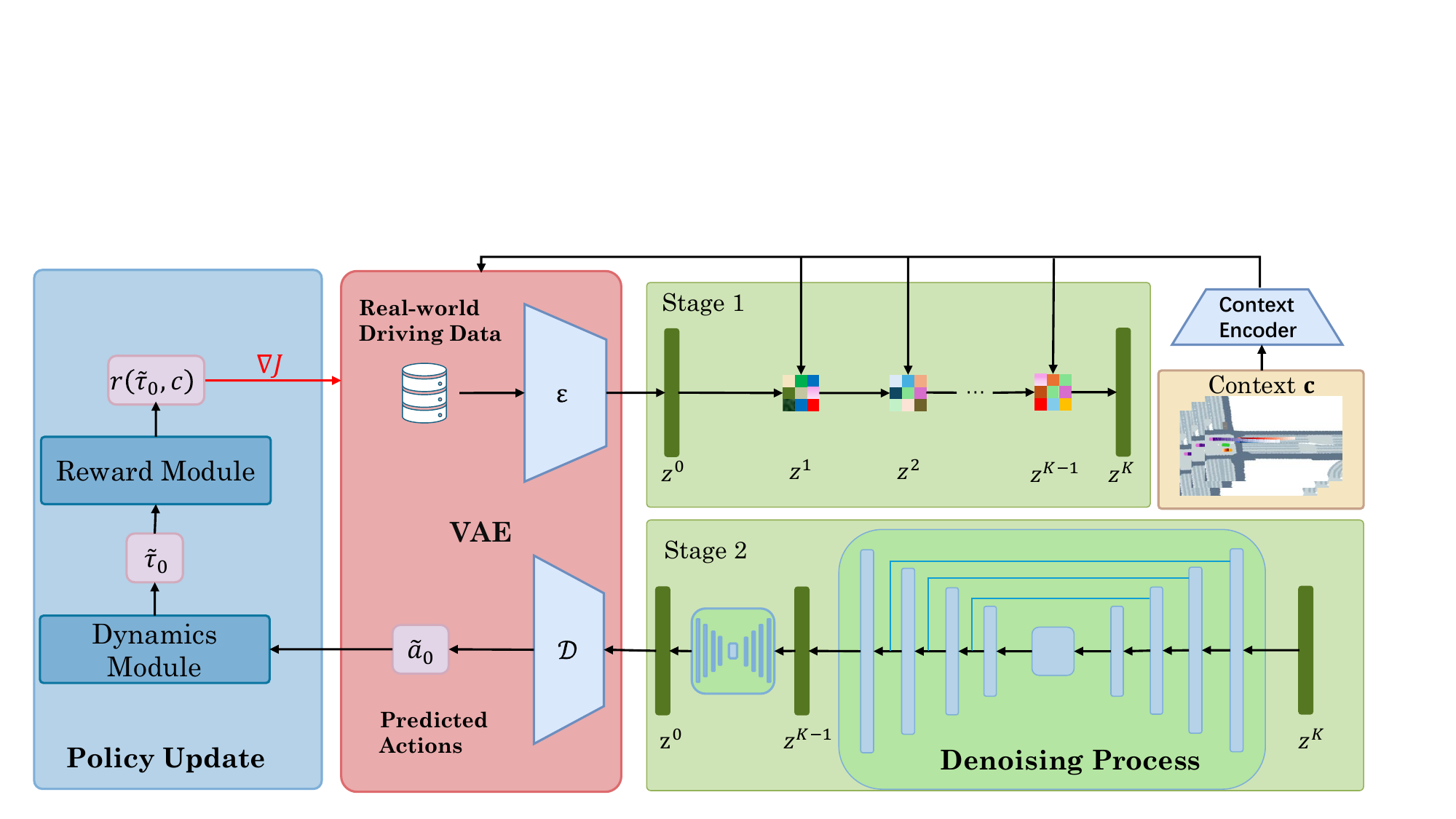}
\caption{Overview of the CLD framework. A VAE is first trained to encode and reconstruct context and trajectories. Afterward, a DM is trained in the latent space to learn the distribution of the original data. Subsequently, the DM iteratively denoises the latent states \(z^k\), which are then reconstructed to generate trajectories. These trajectories are evaluated by a reward module to obtain $\nabla J$ and further optimize the DM, guiding the generation toward controllable and safe traffic scenarios.}
\label{fig:overview}
\end{figure*}

\section{Methodology}
\label{sec:3}
In this section, we present our CLD framework for controllable trajectory generation. We begin by formalizing the problem setting (\cref{sec:problem_formulation}), defining the state and action spaces along with the decision-critical context. Then, we employ a VAE \cite{kingma2013auto} to compress high-dimensional driving trajectories into a compact latent space, which is then used for efficient DM training (\cref{sec:vae_training}). We further detail the DM's operation in the latent space and its training objective (\cref{sec:diffusion_in_latent}). Finally, we introduce our reward-driven MDP formulation that steers the sampling process to generate trajectories meeting user-specified controllability criteria (\cref{sec:reward_guidance}). Our framework is illustrated in \cref{fig:overview}.

\subsection{Problem Formulation}
\label{sec:problem_formulation}

Let state space at time \(t\) defined as $s_t = (x_t,\, y_t,\, v_t,\, \theta_t)$. \(x_t\) and \(y_t\) denote the 2D position, \(v_t\) represents the speed, and \(\theta_t\) is the heading angle. Let \(\mathbf{c}\) denote the decision-critical context, \(\mathbf{c} = (\mathcal{I}, \mathcal{S})\). Here, \(\mathcal{I}\) is a local, agent-centric semantic map providing information such as lane geometry and drivable areas. Meanwhile, \(\mathcal{S}\) aggregates the \(H\) most recent states of both the ego vehicle and its \(M\) neighbors:
\begin{equation}
\mathcal{S}_{t-H:t} = \Bigl\{\, s^{\text{(tgt)}}_{t-H:t},\, s^{\text{(1)}}_{t-H:t},\, \dots, s^{(M)}_{t-H:t}\Bigr\}
\end{equation}
An action \(a_t\) is produced at each time step \(t\), conditioned on the context \(\mathbf{c}\). We define
\begin{equation}
a_t = (\dot{v}_t,\, \dot{\theta}_t;\mathbf{c})
\end{equation}
where \(\dot{v}_t\) denotes longitudinal acceleration and \(\dot{\theta}_t\) the yaw rate. The system state then evolves via a unicycle dynamic function \(f\):
\begin{equation}
\label{eqs:dynamics}
s_{t+1} = f(s_t,\, a_t)
\end{equation}

\noindent\textbf{Trajectory Representation.}  
A driving trajectory is represented as a sequence of states over time T, i.e., 
\begin{equation}
    \mathbf{\tau} = \{s_t,\, s_{t+1},\, \dots,\, s_{t+T}\}
\end{equation}

\noindent\textbf{Objective.}
Our goal is to generate trajectories \(\tau\) that are \emph{realistic} and \emph{stable} while maintaining controllability for customized user demand. To quantify these requirements, we introduce a cost function
\begin{equation}
J(\tau; \mathbf{c}) = \alpha\, \mathrm{Collisions}(\tau;\mathbf{c}) 
\;+\; \beta\, \mathrm{Offroad}(\tau;\mathbf{c})
\end{equation}
where \(\mathrm{Collisions}\) and \(\mathrm{Offroad}\) measure undesirable events. The coefficients \(\alpha, \beta \) balance each term.

We then seek a generative model parameterized by \(\theta\) that produces trajectories \(\tau\) minimizing the expected cost:
\begin{equation}
    \theta^* = \arg\min_{\theta} \; 
\mathbb{E}_{\tau\,\sim\,p_\theta(\cdot \mid \mathbf{c})} \bigl[ J(\tau; \mathbf{c}) \bigr]
\end{equation}
with the state evolution constrained by the dynamics in \cref{eqs:dynamics}.

In the subsequent sections, we detail how we employ a latent DM and reward-driven optimization to achieve both realism and stability under this formulation.

\subsection{VAE Training}
\label{sec:vae_training}
High-dimensional driving scenarios impose significant computational challenges. To mitigate these challenges, the first stage of our framework leverages a VAE \cite{kingma2013auto} to compress driving trajectories into a compact latent space.

\noindent\textbf{Encoder and Decoder.}  
Our VAE model, built on a Long Short-Term Memory (LSTM) \cite{staudemeyer2019understanding} backbone to capture temporal dependencies in sequential data, consists of an encoder \(\mathcal{E}\) and a decoder \(\mathcal{D}\). The LSTM backbone serves as the core architecture for both encoding and decoding. In particular, the encoder processes the driving trajectory \(\tau\) and compresses it into a latent representation \(z\). To incorporate contextual information, we employ a convolutional neural network (CNN) to encode \(\mathbf{c}\) into a compact feature vector, which is then combined with the LSTM's hidden state. Formally, the encoding process can be summarized as:
\begin{equation}
    \mathbf{z} \sim q_{\phi}(\mathbf{z} \mid \tau, \mathbf{c}) = \mathcal{N}\Bigl(z;\, \mu_{\phi}(\tau, \mathbf{c}),\, \sigma_{\phi}^2(\tau, \mathbf{c})\Bigr)
\end{equation}

where \(\mu_{\phi}\) and \(\sigma_{\phi}\) parameterize the latent Gaussian distribution. The LSTM decoder then leverages \(z\) to generate the corresponding actions:
\begin{equation}
    \hat{a}_t \sim p_{\theta}(a_t \mid \mathbf{z})
\end{equation}
which are subsequently used to predict trajectories via a dynamics function.  
We train the VAE by minimizing the loss
\begin{equation}
    \mathcal{L}_{\text{VAE}} = \mathbb{E}_{q_{\phi}(z \mid \tau)}\left[\|\tau - \hat{\tau}\|^2\right] + \beta\, D_{KL}\Bigl(q_{\phi}(\mathbf{z} \mid \tau) \parallel \mathcal{N}(0, I)\Bigr)
\end{equation}
where the reconstruction loss ensures accurate recovery of the input trajectories, and the Kullback-Leibler (KL) divergence regularizes the latent space to approximate a standard Gaussian.

This LSTM-based VAE not only effectively compresses high-dimensional driving trajectories into a compact latent space but also preserves the essential sequential features necessary for subsequent stages of our framework.

\subsection{Diffusion Model in Latent Space}
\label{sec:diffusion_in_latent}

DMs generate ego data distributions by progressively adding noise in a \emph{forward diffusion process} and then iteratively removing that noise in a \emph{denoising process} \cite{sohl2015deep}. We apply this paradigm in the latent space of our VAE to efficiently produce realistic and stable driving trajectories.

\noindent\textbf{Forward Diffusion Process.}
We define the forward diffusion process as a Markov chain that progressively corrupts an initial latent code \( \mathbf{z}_0 \) by adding Gaussian noise over \( K \) steps. Formally, for each step \( k = 1, \dots, K \):
\begin{equation}
q\bigl(\mathbf{z}^k \mid \mathbf{z}^{k-1}\bigr) := 
\mathcal{N}\Bigl(\mathbf{z}^k; \sqrt{1-\beta_k}\,\mathbf{z}^{k-1},\, \beta_k\,\mathbf{I}\Bigr)
\end{equation}
where \(\{\beta_k\}_{k=1}^{K}\) is a pre-defined variance schedule. After \(K\) steps, the latent code is nearly indistinguishable from a standard Gaussian, i.e., \(\mathbf{z}^K \sim \mathcal{N}(\mathbf{0},\mathbf{I})\).

\noindent\textbf{Training Objective.}
During training, the DM learns to predict the noise added in the forward process, effectively recovering the original latent code. The simplified loss function is:
\begin{equation}
    \mathcal{L}_{\mathrm{DM}} 
= \mathbb{E}_{\mathbf{z},\, \epsilon \sim \mathcal{N}(0,1),\, ,\, \mathbf{c}}
\Bigl[
\|\epsilon - \epsilon_{\theta}(\mathbf{z}^k,\, k)\|^2
\Bigr]
\end{equation}
$\epsilon$ is the Gaussian noise sampled from \(\mathcal{N}(\mathbf{0}, \mathbf{I})\).

\subsection{Reward-Driven MDP}
\label{sec:reward_guidance}      

\noindent\textbf{Denoising Process.}
To generate new latent representations, a sampling process that denoises the noisy latent states is executed by the learned DM. Conditioned on an auxiliary context \(\mathbf{c}\) derived from the VAE, the sampling process is formulated as:
\begin{equation}
p_\theta\bigl(\mathbf{z}^{0:K} \mid \mathbf{c}\bigr) 
:= p\bigl(\mathbf{z}^K\bigr) 
\prod_{k=K}^{1} p_\theta\bigl(\mathbf{z}^{k-1}\mid \mathbf{z}^k, \mathbf{c}\bigr)
\end{equation}
with each denoising step modeled as:
\begin{equation}
    p_\theta\bigl(\mathbf{z}^{k-1}\mid \mathbf{z}^k, \mathbf{c}\bigr) 
    := \mathcal{N}\Bigl(\mathbf{z}^{k-1};  
    \boldsymbol{\mu_\theta}\bigl(\mathbf{z}^k, k, \mathbf{c}\bigr),  
    \boldsymbol{\Sigma_\theta}\bigl(\mathbf{z}^k, k, \mathbf{c}\bigr)\Bigr)
\end{equation}

We model the denoising diffusion process as a multi-step MDP \cite{black2023training}, where both the policy and reward are defined as follows. 
\begin{equation}
\begin{split}
s_t &\triangleq (c,\, k,\, \mathbf{z}^k), \quad \pi(a \mid s) = p_\theta(\mathbf{z}^{k-1} \mid \mathbf{z}^k,\, c) \\
a_t &\triangleq \mathbf{z}^{k-1}, \quad R(s,\, a) = 
\begin{cases}
r(\tau_0,\, c), & \text{if } k = 0\\[1mm]
0 & \text{otherwise}
\end{cases}
\end{split}
\end{equation}

By formulating the denoising diffusion process as an MDP, we generate sample trajectories from the diffusion process and compute each trajectory's accumulated reward, which is then summed and averaged over these samples \cite{black2023training}. This integration facilitates a more flexible and principled approach to policy optimization.

\noindent\textbf{Multiple Steps Policy Optimization.}
We adopt an importance sampling estimator \cite{kakade2002approximately} that allows us to reuse existing data for multiple optimization steps, thereby improving sample efficiency.
\begin{equation}
\begin{split}
\nabla_\theta J_{\text{DDRL}} = \mathbb{E}\Biggl[\, &\sum_{k=0}^K 
\frac{p_\theta(\mathbf{z}^{k-1}\mid \mathbf{z}^k, c)}
{p_{\theta_{\text{old}}}(\mathbf{z}^{k-1}\mid \mathbf{z}^k, c)} \\
&\quad \times \nabla_\theta \log p_\theta(\mathbf{z}^{k-1}\mid \mathbf{z}^k, c)\, r(\tau_0, c)
\Biggr]
\end{split}
\end{equation}
where the expectation is taken over denoising trajectories generated by the old policy \(p_{\theta_{\text{old}}}\). By applying importance sampling \cite{kakade2002approximately}, our estimator reuses the collected data, thereby improving sample efficiency.

The overall guided training procedure is summarized in \cref{alg:reward_guided_sampling}.

\begin{algorithm}[H]
\caption{Guided Diffusion Model Parameter Update}
\label{alg:reward_guided_sampling}
\begin{algorithmic}[1]
\Require Encoder $\mathcal{E}$, conditional diffusion model $\mu_{\theta}$, dynamics function $f$, decoder $\mathcal{D}$, guidance function $J$, covariance matrices $\{\Sigma_k\}_{k=1}^{K}$, number of diffusion steps $K$, execution horizon $l$, and learning rate $\eta$.
\State \textbf{Input:} Current state $s_0$ and context $c$
\State \textbf{Latent Initialization:} Sample $\mathbf{z}^K \sim \mathcal{N}(\mathbf{0}, \mathbf{I})$
\For{$k = K$ \dots $1$}
    \State $\mu \gets \mu_{\theta}(\mathbf{z}^k, k, c)$
    \State Sample $\mathbf{z}^{k-1} \sim \mathcal{N}(\mu, \Sigma_k)$
\EndFor
\State \textbf{Decoding:} Compute action $a \gets \mathcal{D}(z^0)$
\State Compute state $\mathcal{\tau}_0 \gets f(s_0, a)$

\State \textbf{Scoring:} Compute reward $r \triangleq r(\mathcal{\tau}_0, c)$

\State\textbf{Parameter Update:} Update the diffusion model parameters:
\[
\theta \gets \theta + \eta\, \nabla_\theta J_{\text{DDRL}}
\]

\end{algorithmic}
\end{algorithm}

\begin{figure*}[htbp]
\centering
\includegraphics[width=\textwidth]{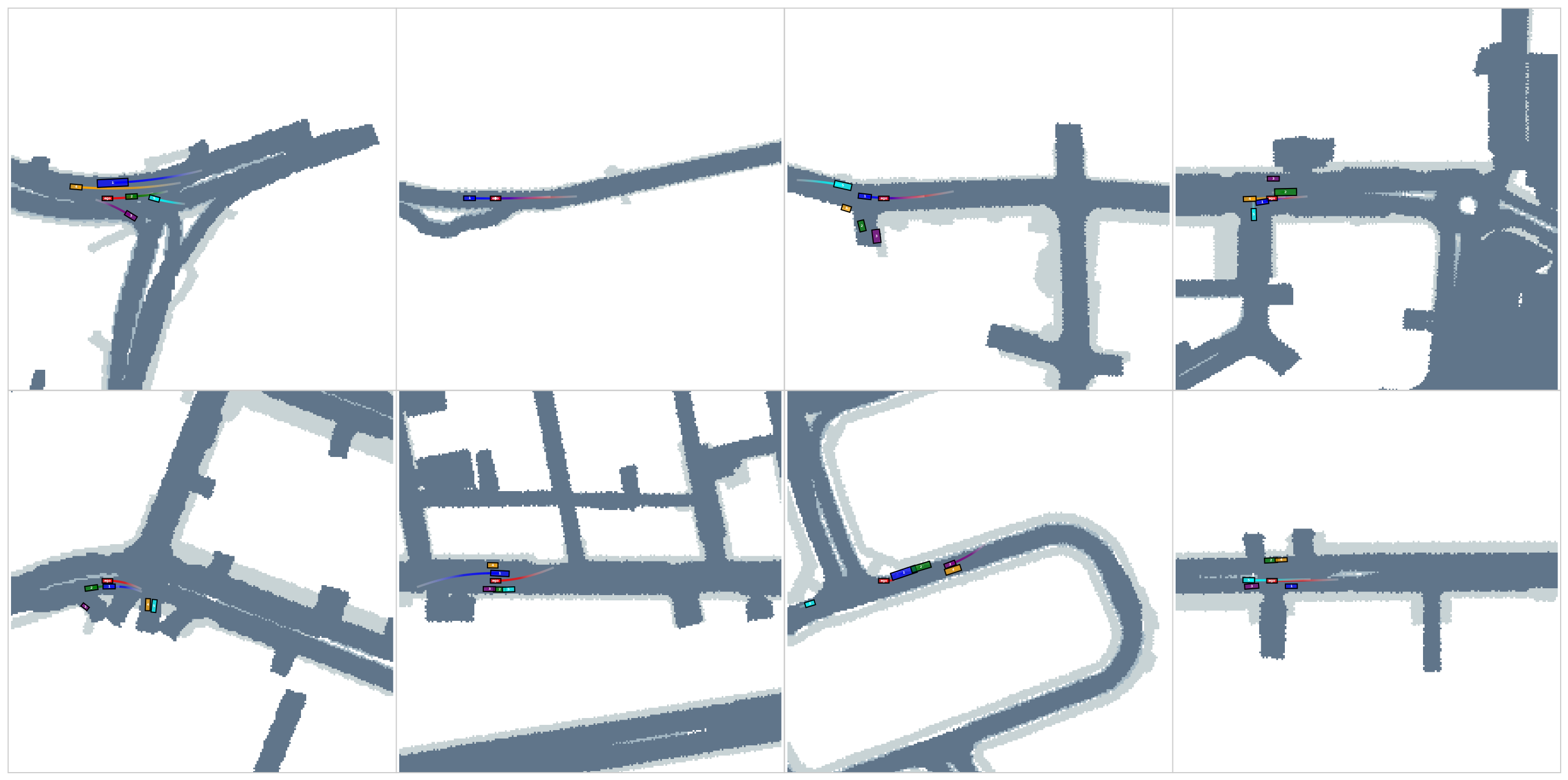}
\caption{Generated trajectories in various traffic simulations by CLD. In each scenario, the red line indicates the trajectory of the ego agent, while other colors represent surrounding agents. The generated trajectories in complex intersections show collision-free and traffic rules compliant behaviors, demonstrating the robust and adaptive performance of our approach across diverse driving conditions.}

\label{fig:output}
\end{figure*}

\section{Experiments}
\label{sec:4}
We conduct an experiment to evaluate our framework for controllable traffic simulation. We detail the experimental setup (\cref{sec:experimental setup}), including the simulation environment, data, and evaluation metrics. We then analyze the experimental results by comparing our approach with several baseline methods, while highlighting both its strengths and limitations (\cref{sec:results}). Finally, we conduct an ablation study to demonstrate that incorporating an appropriate reward to guide the diffusion process is critical for achieving controllability. Our ablation results confirm that reward-guided diffusion significantly improves the controllability of the generated trajectories, as evidenced by reduced failure rates, while maintaining high levels of realism (\cref{sec:ablation study}).

\subsection{Experimental Setup}
\label{sec:experimental setup}
Our approach is evaluated on the nuScenes dataset \cite{caesar2020nuscenes},  a large-scale multimodal dataset that encompasses over 5.5 hours of driving data collected in urban environments across multiple cities. This dataset offers accurate vehicle trajectories, sensor readings, and detailed semantic maps, making it an ideal benchmark for testing AV systems. Our evaluation focuses on two main aspects: \textit{realism} and \textit{stability}.

\textit{Realism} is assessed by comparing key driving dynamics: longitudinal acceleration, lateral acceleration, and jerk \cite{xu2023bits} between generated trajectories and ground truth data. Specifically, we compute the Wasserstein distance between the distributions of key driving metrics derived from the ground truth dataset and those from the generated trajectories. We then define an aggregated metric, denoted as \emph{real}, as the mean Wasserstein distance across these three distributions.

\textit{Stability} is evaluated by the failure rate, defined as the percentage of scenarios in which the ego trajectory either collides with obstacles or other agents or deviates from the drivable area.

\noindent\textbf{Reward Design.} To enforce the no-collision criterion, we compute the Euclidean distance between the predicted trajectory and the positions of all other agents over \(t\) time steps. If the distance at any time step falls below a predefined threshold (indicating a collision), the entire trajectory is assigned a reward of \(-1\); otherwise, it receives a reward of \(0\).
We evaluate off-road behavior by verifying that the predicted trajectory remains within the designated drivable area at all time steps. If any part of the trajectory is located outside this region, a reward of \(-1\) is assigned to the entire trajectory; otherwise, the trajectory receives a reward of \(0\).

\noindent\textbf{Baselines.}
For comparison, we use four baseline methods. \textit{SimNet} \cite{bergamini2021simnet}: a deterministic behavior-cloning model, along with its optimized variant (SimNet+opt) that applies optimization on the output action trajectory. \textit{TrafficSim} \cite{suo2021trafficsim}: a conditioned VAE-based trajectory generation method evaluated in two forms—one variant employing filtration with our loss function, and another (TrafficSim+opt) that further incorporates latent space optimization. \textit{BITS} \cite{xu2023bits}: a bi-level imitation learning model, with an additional variant (BITS+opt) that applies optimization on the output action trajectory.
\textit{CTG w/o f+g}: a baseline corresponding to CTG without filtration and guidance, which incorporates guided sampling in a DM.

\noindent\textbf{Closed-loop Traffic Simulation.}
We evaluate our model using closed-loop traffic simulations, where each agent iteratively generates short-horizon trajectory predictions and executes corresponding actions. Simulations are initialized using ground-truth states from the nuScenes dataset and run for 20 seconds with a re-planning frequency of 2 Hz. This closed-loop setup ensures predictions dynamically adapt to evolving environmental conditions, accurately reflecting realistic driving scenarios.

\subsection{Results}
\label{sec:results}
We evaluate the overall performance of our controllable trajectory generation framework. Unlike previous works that assess performance on individual rule-based settings (e.g., single or multiple rule evaluations) and perform ablation studies, our evaluation focuses on the holistic quality of the generated trajectories, measured in terms of realism and stability. 

\Cref{fig:output} illustrates the trajectories generated by CLD in various complex scenarios. The generated trajectories of the ego agent (as illustrated in red) consistently avoid collisions and strictly adhere to traffic rules, underscoring the robustness and adaptability of our approach under diverse driving conditions.

\cref{table:1} compares our approach against the baselines under the evaluation metrics outlined in \cref{sec:experimental setup}, whereas the baselines' performance is from \cite{zhong2023guided}.
In the experiments, our approach demonstrates impressive performance on both collision avoidance and drivable area following. For the no-collision task, our method achieves a failure rate of \(0.226\) — significantly lower than those of the baselines — indicating that the generated trajectories effectively avoid collisions. Similarly, for the no off-road task, our approach yields a realism score of \(0.655\) and a failure rate of \(0.136\), demonstrating that the trajectories consistently remain within the designated drivable area and exhibit strong fidelity to real-world conditions. Together, these results confirm that our framework robustly generates scenarios that meet critical safety and performance criteria.

\setlength{\tabcolsep}{12pt}
\begin{table}[t]
\centering
\caption{Comparison across multiple tasks, each with rule-based and real metrics.}
\label{table:1}

\begin{tabular}{l|cc|cc}
\hline
& \multicolumn{2}{c|}{\textbf{no collision}}
& \multicolumn{2}{c}{\textbf{no off-road}} \\
\hline
\textbf{Method} & \textbf{real} & \textbf{fail} & \textbf{real} & \textbf{fail} \\
\hline
SimNet       & 0.898 & 0.353  & 0.900 & 0.353 \\
SimNet+opt   & 1.149 & 0.398 & 1.242 & 0.416  \\
TrafficSim   & 1.542 & 0.416 & 1.564 & 0.401 \\
TrafficSim++ & 1.063 & 0.265 & 1.836 & 0.413 \\
BITS         & 1.220 & 0.314 & 1.131 & 0.296 \\
BITS+opt     & 1.617 & 0.354 & 1.261 & 0.343 \\
CTG (w/o f+g)& 0.396 & 0.301 & 0.396 & 0.301 \\
CTG          & 0.596 & 0.271 & 0.501 & 0.455 \\
\textbf{CLD (w/RL)}& \textbf{0.655} & \textbf{0.226} &\textbf{0.655}  &\textbf{0.136}  \\
\textbf{CLD (w/o RL)}& \textbf{0.350} & \textbf{0.483} &\textbf{0.350}  &\textbf{0.450}  \\
\hline
\end{tabular}
\end{table}

\subsection{Ablation Study}
\label{sec:ablation study}
To validate the effectiveness of our training-guided mechanism, we conducted an ablation study comparing the full model (with reward guidance) to a variant without it. Notably, when reward guidance is removed, the realism metric decreases to \(0.350\) for both tasks — indicating that the generated trajectories more closely resemble the ground truth. However, this improvement in realism comes at a significant cost: the failure rates increase to \(0.483\) and \(0.450\), respectively, demonstrating reduced controllability. These results confirm that our approach is essential for achieving a balanced trade-off between realism and controllability.

\section{Conclusion}\label{sec:conclusion}
In this paper, we presented CLD, a framework capable of generating traffic scenarios with enhanced controllability across various tasks. Central to our approach is a guided training strategy, in which a simple reward mechanism steers the generative process to achieve high realism and stability in the output trajectories. In addition, the guided training enables controllable scenario generation. Nonetheless, many avenues remain for further exploration. In future work, we plan to extend our framework to generate an even wider range of safety-critical scenarios by adapting the reward module and incorporating additional types of agents, thereby further broadening its applicability in AV safety evaluation.

\addtolength{\textheight}{-3cm}

\bibliographystyle{IEEEtran}  
\bibliography{reference}  

\end{document}